# Desarrollo de un Algoritmo para el Control de Desplazamiento de una Plataforma Robótica, Mediante el Procesamiento Digital de Imágenes[1]

# Development of an Algorithm for Controlling Movement of a Robotic Platform, by Digital Image Processing

A. Huérfano, H. Numpaque y L. Díaz

*Resumen* - El trabajo que aquí se presenta es el desarrollo de un algoritmo que permite procesar imágenes de manera digital con el objeto de controlar el desplazamiento de una plataforma móvil robótica en un determinado entorno. La plataforma está identificada con un color especifico y el entorno de desplazamiento de la plataforma tiene obstáculos identificados con diferentes colores, para ambos casos se trabajó con la escala de colores RGB.

Para lograr el control del desplazamiento de la plataforma robótica, el algoritmo se desarrolló en lenguaje de programación C, y se emplearon las librerías OpenCV para hacer el procesamiento de imágenes capturadas por una cámara de video sobre la plataforma Dev-C++.

La cámara de video previamente fue calibrada empleando la técnica ZHANG de donde se obtuvieron los parámetros de longitud focal, centro focal e inclinación de pixel. En el algoritmo se desarrollaron análisis de histogramas y segmentación de la imagen, permitiendo determinar con exactitud la posición relativa de la plataforma con respecto a los obstáculos y la estrategia de movimiento a seguir.

*Palabras Clave* - Procesamiento Digital de Imágenes, Histogramas, Segmentación, Posición Relativa, OpenCV.

*Abstract* - The following work shows an algorithm that can process images digitally with the goal of control the movement of a mobile robotic platform in a certain environment. The platform is identified with a specific color, and displacement environment of the platform shift has identified obstacles with different colors, for both cases it worked with the RGB color scale.

To obtain the control's movement of the robotic platform, the algorithm was developed in C programming language, and used the Open CV libraries for processing images captured by a video camera on the Dev-platform C + +.
The video camera was previously calibrated using ZHANG technique where parameters were obtained focal length and tilt focal pixel. In the algorithm histogram analysis and segmentation of the image were developed, allowing to determine exactly the relative position of the platform with respect to the obstacles and movement strategy to follow.

I. Introducción

La robótica ha venido desempeñando un papel fundamental en la automatización y optimización de procesos de producción a nivel industrial, donde los algoritmos de visión artificial juegan un rol fundamental en el desempeño de dichas estructuras en este entorno.

Una de las necesidades más importantes es la detección de determinado elemento mediante su forma o color para así llevarlo de un lugar a otro, esto con la finalidad de agilizar procesos de almacenamiento y desplazamiento para economizar recursos en mano de obra. Ante esto se estableció la necesidad de generar un mecanismo que definiera una imagen como una función de dos dimensiones (X, Y), cada vez que las coordenadas tomen valores finitos, la imagen puede convertirse en una matriz y así generar un

20

procesamiento digital de imágenes [1] (Análisis generado a partir de imágenes tomadas previamente o en tiempo real, con una cámara).

El presente trabajo muestra el avance de un algoritmo que se está desarrollando en el Semillero de Investigación (SIRIA (Semillero de Investigación de Robótica Industrial Aplicada) de la Universidad de Cundinamarca, dicho algoritmo es el resultado del estudio bibliográfico especializado en procesamiento de imágenes OpenCV y la plataforma de programación Dev-C++, con la finalidad de generar el control de desplazamiento de una plataforma robótica que centra su funcionalidad en evadir obstáculos y desplazar objetos. Son dos las características principales para el éxito de la plataforma; en primer lugar, la escala de color escogida para trabajar que es RGB; y segundo, la correcta calibración de una cámara con resolución VGA que captura las imágenes.

## II. Métodos y materiales

Una de las partes más importantes en el desarrollo del proyecto es la escala de color empleada en el trabajo; el modelo de color RGB fue escogido ya que es uno de los modelos que está orientado al Hardware [2], este aditivo se compone de tres colores (rojo, verde, azul) y se suman para producir otros diferentes. Las pantallas electrónicas lo usan, significando con esto que los colores no son absolutos, sino que dependen de la sensibilidad y la configuración de cada dispositivo [3].

Otra importante característica es el espacio en el que se va a desplazar la plataforma, este será sobre una pista uniforme y unicolor, la plataforma robótica será identificada dentro del algoritmo como el color *verde*, el objetivo hasta donde tiene que llegar estará identificado como el color *rojo* y todo lo que para él será un obstáculo *azul*. La Fig.1, muestra el esquema físico del entorno donde será desplazada la plataforma.

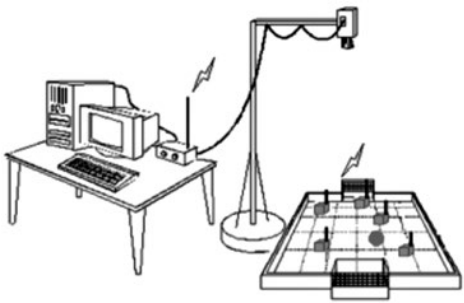

Fig. 1. Esquema de la estructura física en la que se llevará a cabo la implementación del proyecto [4].

## III. Calibración de la cámara

Dentro del proceso desarrollado, la calibración de la cámara es uno de los estamentos más importantes, ya que establece bajo qué condiciones focales se está trabajando, el método utilizado para ello es el método de ZHANG a través de MATLAB, dicho método de calibración está basado en la observación de una plantilla plana desde varias posiciones. Mediante este paso se obtienen los parámetros tanto intrínsecos (la distancia focal, el punto principal y el centro óptico) como extrínsecos (orientación del cuadro de referencia de la cámara con respecto al mundo real), es decir, dan la orientación externa de la cámara fácilmente [5], a partir de una plantilla plana en la cual no es necesario conocer la posición de los puntos de interés, tampoco es necesario conocer las posiciones de la cámara desde donde se han tomado las imágenes de la misma [6].

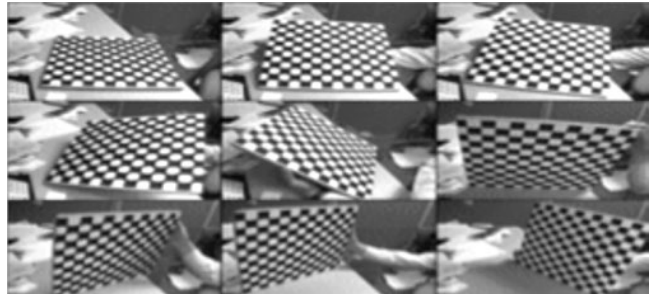

Fig. 2. Esquema básico de la forma en que deben ser tomadas las imágenes.

La Fig.2 y Fig.3 muestran respectivamente la plantilla que se utiliza para tomar las fotos que se utilizarán en el proceso de calibración y el diagrama de flujo empleado para calibrar la cámara. Matlab posee un ToolBox que permite hacer con mayor facilidad este proceso. La calibración consiste básicamente en hacer con la cámara un determinado número de tomas (entre 20 y 25) desde todos los ángulos de la plantilla (Ajedrez de 14x13, 3x3) y con las mismas condiciones lumínicas. Es de vital importancia que ninguna de las imágenes quede recortada y que el Matlab que posea el ordenador tenga incluido el ToolBox de calibración [7, 8].

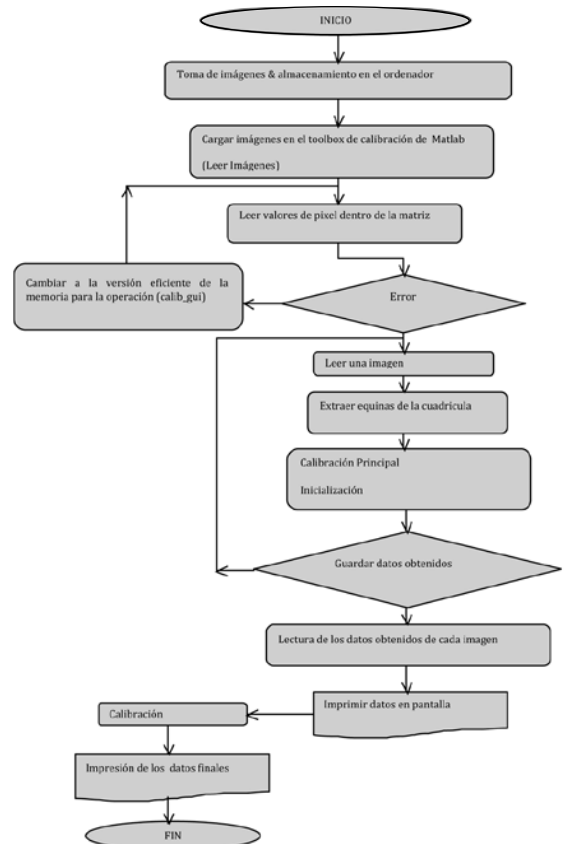

Fig. 3. Diagrama de flujo empleado para calibrar la cámara.

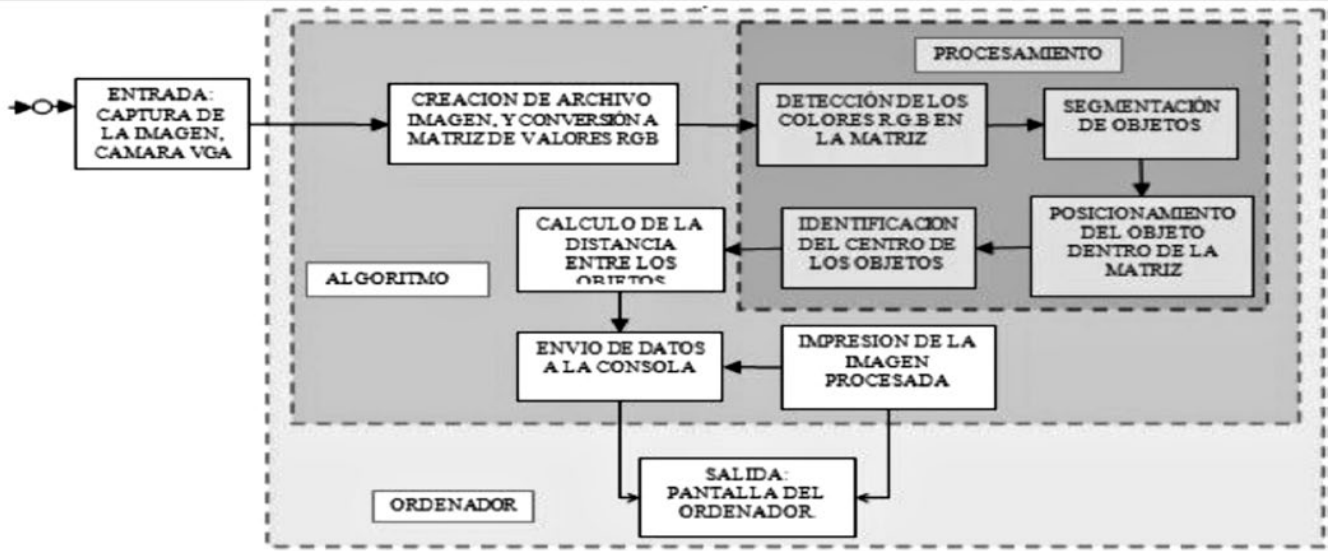

Fig. 4. Diagrama de Diagrama de bloques del proceso desarrollado.

La calibración de la cámara es de vital importancia para el desarrollo del proceso, ya que mediante los datos que sean arrojados al hacerla, se pueden establecer parámetros como la inclinación del pixel y obtener la relación entre centímetros-pixel a la hora de la implementación mostrada en la Fig.1.

## IV. Metodología de desarrollo del algoritmo en OpenCv

El algoritmo que permite hacer el desplazamiento de una plataforma robótica mediante el procesamiento digital de imágenes está generado sobre el software de programación Dev-C++ y afianzado en las librerías especializadas en procesamiento de imagen OpenCv [9]. El diagrama de bloques de la Fig.4, permite ver de forma general, el planteamiento y desarrollo del proceso. Dicho algoritmo, tiene la capacidad de hacer la detección de los colores y formas de un espacio determinado, para así generar las distancias halladas entre ellos y poderlas presentar en consola. La Fig. 5 muestra el diagrama de flujo del algoritmo.

## V. Análisis de resultados

El inicio de todo el proyecto radica en la escala de color a utilizar, que para el caso es la escala de color RGB (Red Green Blue), fue seleccionada por ser la más utilizada para aplicaciones llevadas al Hardware [2].

La elección de los colores a detectar se establece tomando en cuenta las necesidades que se tenían y la forma más ágil y eficiente de satisfacerlas, es claro que por el momento hay que detectar tres puntos (Plataforma robótica, Objetivo y Obstáculo), así que se decide escoger los colores Rojo (El objetivo de la plataforma robótica), Verde (Plataforma robótica) y Azul (Obstáculos de la plataforma robótica); que estos son los principales dentro de esta gama de colores.

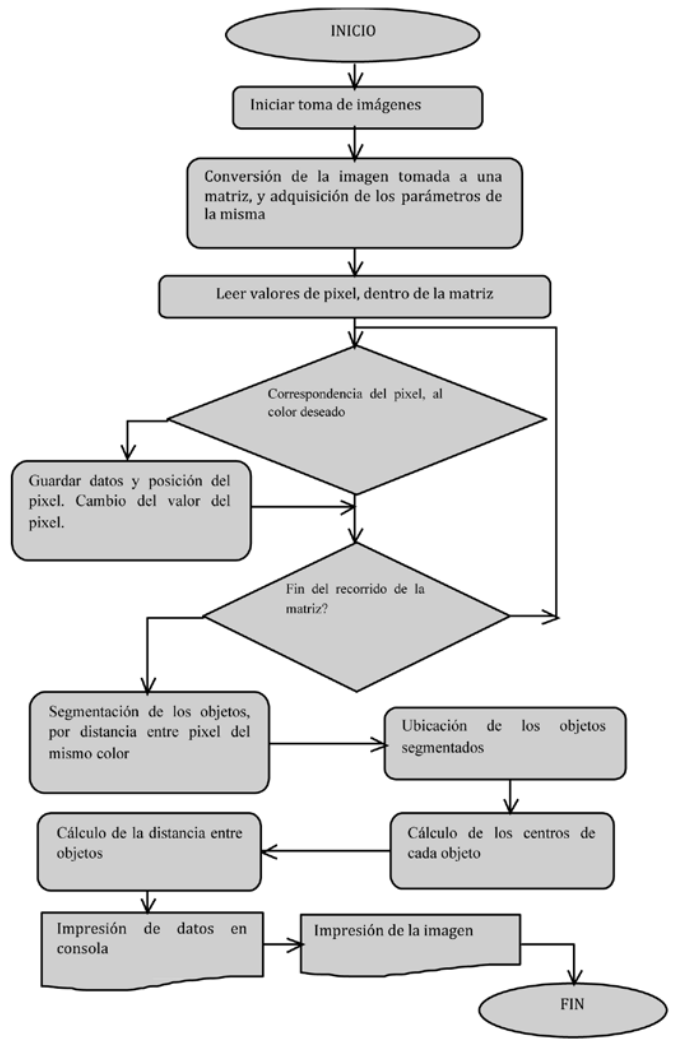

Fig. 5. Diagrama de flujo del algoritmo.

Dejando claro lo anterior y partiendo de esto, se procede a hacer la captura de la imagen, para que sea convertida en matriz dentro del algoritmo y escanear detalladamente toda la imagen ya que es recorrida pixel a pixel. Así se obtiene la información de cada pixel para determinar si hay o no un objeto del color establecido.

Teniendo los objetos de dichos colores plenamente localizados se halla el centro de los mismos, esto se logra mediante un promedio que se realiza de la cantidad de los pixeles ocupados por la imagen tanto en el eje X como en el eje Y, con esto se llega al centro del objeto dentro de la imagen. Haciendo uso de la aritmética básica (operación entre vectores y teorema de Pitágoras), se hallan las distancias entre los objetos.

La Fig. 5 muestra el diagrama de flujo detallado del algoritmo realizado bajo la anterior descripción.

Con esto se muestra en el ordenador la ventana de captura, una venta con la imagen procesada, la consola y un archivo .txt (Este con el fin de que sea usado posteriormente en un aplicativo, para que sea enviado al Microcontrolador, el cual gobernará la plataforma robótica) que muestra los datos adquiridos. Los anteriores resultados se muestran en la Fig. 6.

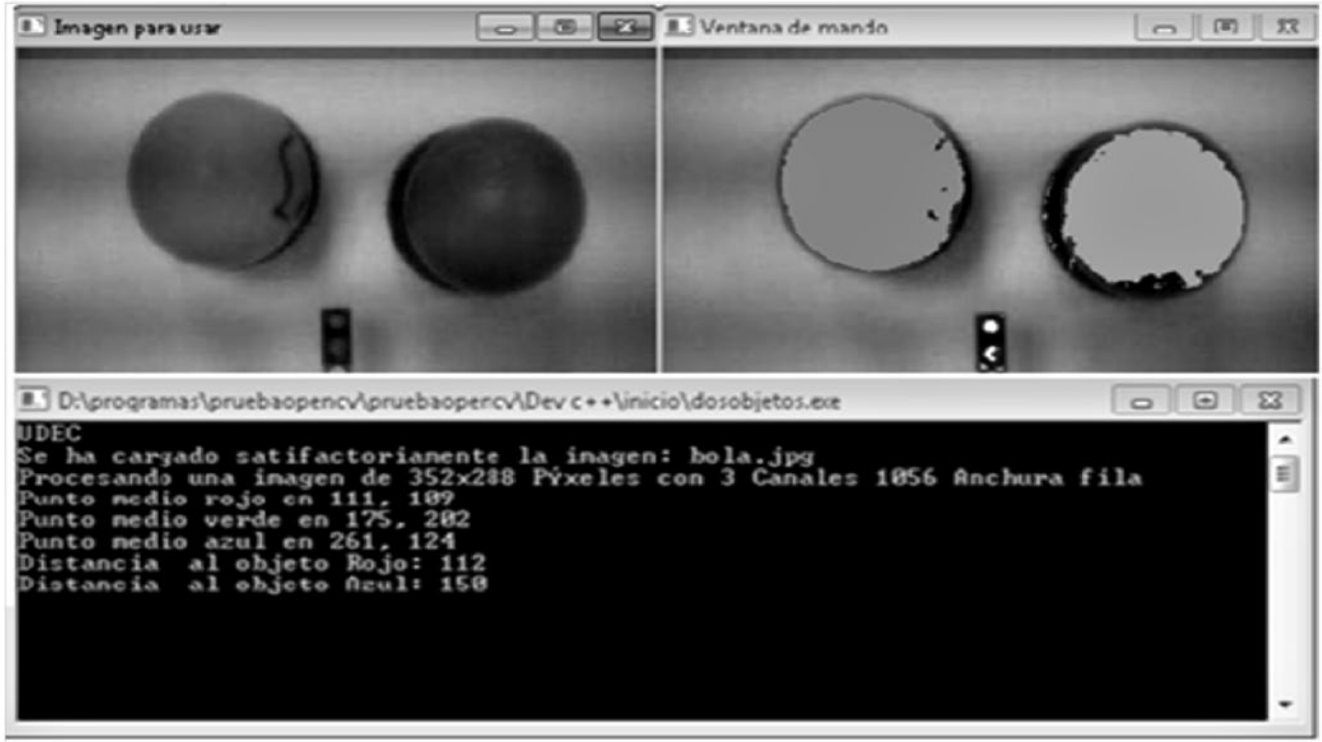

Fig. 6. Resultados del desarrollo del algoritmo en OpenCV que determina el centro de los objetos, y la distancia en pixeles del centro del objeto verde (Plataforma robótica) al centro de los objetos de color rojo (Objetivo de la plataforma robótica) y azul (obstáculos).

## VI. Conclusiones

Mediante el procesamiento digital de imágenes se pueden generar diversas cantidades de aplicativos que en la actualidad son de amplio uso, esta técnica aplicada a la robótica logra ser de gran utilidad en procesos de desplazamiento, almacenamiento a nivel industrial, y reconocimiento de objetos.

A través del uso de librerías OpenCV, para la generación de algoritmos robustos, se pueden realizar de manera rápida y satisfactoria la detección de colores RGB dentro de un ambiente en movimiento, permitiendo umbrales de cambios atmosféricos a la hora de la segmentación, para que de esa forma el algoritmo sea menos sensible a las condiciones lumínicas.

A partir de la detección de objetos mediante el procesamiento digital de imágenes en ambientes variables, se crea una concepción mucho más dinámica y completa del mismo en comparación con el uso de sensores (ópticos, ultrasonido, etc), lo cual le permite tener una gran ventaja con respecto al uso e implementación en una plataforma robótica, dotada de un medio de comunicación bidireccional con un ordenador.

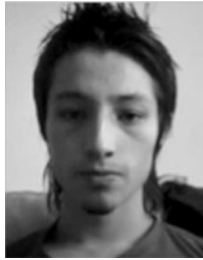

**Andrés Huérfano** (M'1992- ) nació en Colombia, el 12 de Junio de 1992. Cursa actualmente novenos semestre de ingeniería electrónica en la Universidad de Cundinamarca. Forma parte del semillero de investigación SIRIA (Semillero de Investigación de Robótica Industrial Aplicada) en cual esta cobijado por el grupo de investigación GITEINCO (Grupo de Investigación en Tecnologías de la Información y las Comunicaciones). Dentro de este hace parte del proyecto de 'Desarrollo de un algoritmo para el control de desplazamiento de una plataforma robótica, mediante el procesamiento digital de imágenes' el cual está actualmente en ejecución.

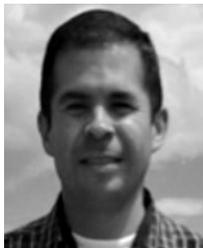

**Humberto Numpaque** (M'1979- ) nació en Colombia, el 9 de septiembre de 1979. Ingeniero electrónico de la Universidad Distrital Francisco José de Caldas 2005, Magister en Ingeniería de Control Industrial 2012. Actualmente es profesor del Departamento de Ingeniería Electrónica, Universidad de Cundinamarca, e investigador grupo GITEINCO (Grupo de Investigación en Tecnologías de la Información y las Comunicaciones) clase D Colciencias. Sus principales líneas de investigación son: electrónica de potencia, sistemas de control, y robótica.

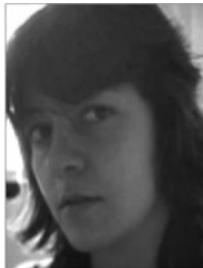

**Lorena Díaz** (F'1992- ) nació en Colombia, el 5 de Agosto de 1992. Cursa actualmente novenos semestre de ingeniería electrónica en la Universidad de Cundinamarca. Forma parte del semillero de investigación SIRIA (Semillero de Investigación de Robótica Industrial Aplicada) en cual esta cobijado por el grupo de investigación GITEINCO (Grupo de Investigación en Tecnologías de la Información y las Comunicaciones). Dentro de este hace parte del proyecto de 'Desarrollo de un algoritmo para el control de desplazamiento de una plataforma robótica, mediante el procesamiento digital de imágenes' el cual está actualmente en ejecución.